\pdfoutput=1

\documentclass[11pt]{article}

\usepackage{acl}
\usepackage{graphicx}
\usepackage{dsfont}
\usepackage{bbold}
\usepackage{booktabs}

\usepackage{times}
\usepackage{latexsym}
\usepackage{multirow}
\usepackage{natbib}

\usepackage[T1]{fontenc}
\usepackage{tikz}
\usepackage{graphicx}
\usepackage{svg}

\usepackage[export]{adjustbox}

\graphicspath{ {./figs/} }

\usepackage[utf8]{inputenc}

\usepackage{microtype}
\usepackage{amsmath}
\usepackage{multirow}
\usepackage{makecell}

\usepackage{xcolor}
\usepackage[inline]{enumitem}

\newcommand\breakdataset{\textsc{Break}}
\newcommand\smcalflow{\textsc{SMCalFlow}}
\newcommand\mtop{\textsc{MTop}}
\newcommand\ourmethod{EPR}

\newcommand{\hyperblue}[2]{\href{#1}{\color{blue} #2}}

\newcommand\eee{e}

\newcommand\sD{\mathcal{D}}
\newcommand\sP{\mathcal{P}}

\newcommand\sE{\mathcal{E}}

\title{Learning To Retrieve Prompts for In-Context Learning}

\author{Ohad Rubin ~~~~~~~ Jonathan Herzig ~~~~~~~
Jonathan Berant \\
The Blavatnik School of Computer Science, Tel Aviv University \\
 \small{\texttt{\{ohad.rubin,jonathan.herzig,joberant\}@cs.tau.ac.il}}
}

\begin{document}
\maketitle


\begin{abstract}
\emph{In-context learning} is a recent paradigm in natural language understanding, where a large pre-trained language model (LM) observes a test instance and a few training examples as its input, and directly decodes the output without any update to its parameters. However, performance has been shown to strongly depend on the selected training examples (termed \emph{prompts}). In this work, we propose an efficient method for retrieving prompts for in-context learning using annotated data and an LM. Given an input-output pair, we estimate the probability of the output given the input and a candidate training example as the prompt, and label training examples as positive or negative based on this probability. We then train an efficient dense retriever from this data, which is used to retrieve training examples as prompts at test time. We evaluate our approach on three sequence-to-sequence tasks where language utterances are mapped to meaning representations, and find that it substantially outperforms prior work and multiple baselines across the board.

\end{abstract}

\section{Introduction}
The striking language skills and world knowledge embedded in large pre-trained language models (LMs) \citep{devlin-etal-2019-bert,petroni-etal-2019-language,raffel2020t5,brown2020language} have recently led to \emph{in-context learning}, a new paradigm in natural language understanding. Under this paradigm, a language model is
given a \emph{prompt}, which typically contains a few training examples, as well as a test instance as input, and generates the output for the test instance directly, without any update to its parameters. This approach was first introduced in GPT-3 \cite{brown2020language}, but has quickly spread to other LMs \cite{lieber2021jurassic,du2021glam,rae2021scaling}.

\begin{figure}
    \noindent

    \includegraphics[width=0.50\textwidth,left]{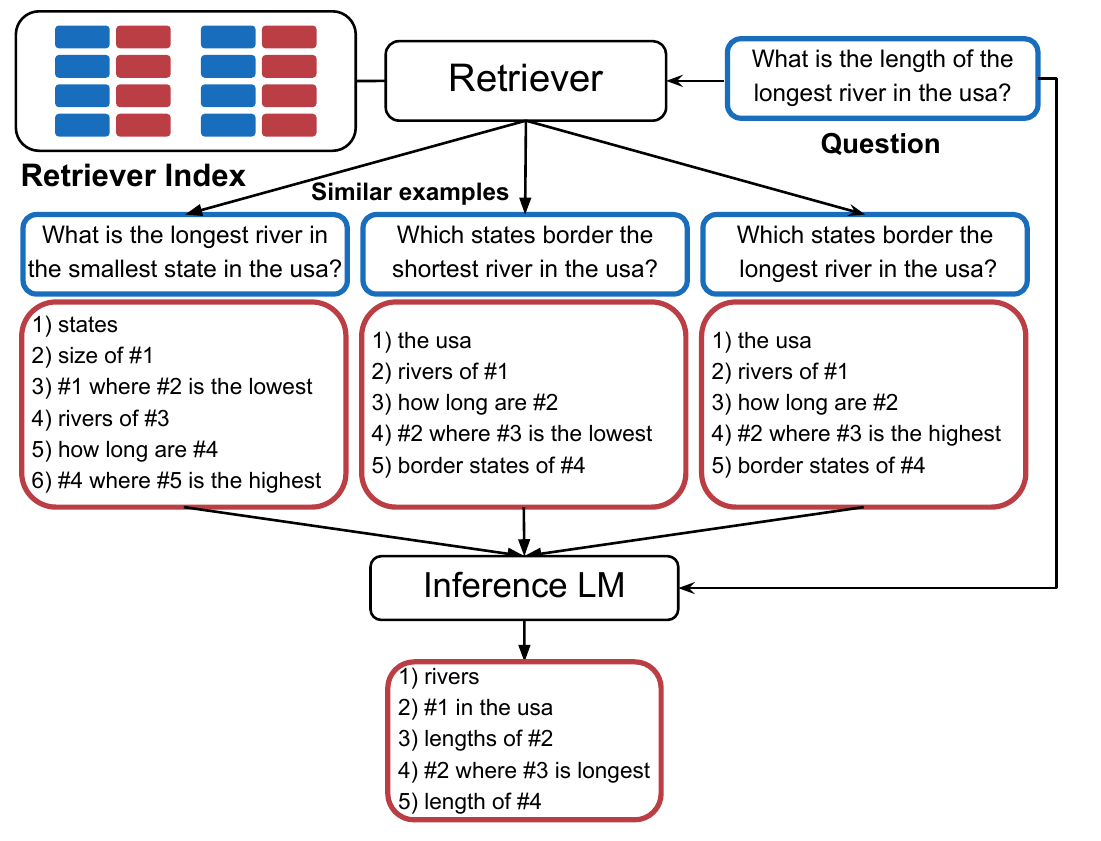}

    \caption{An overview of prompt retrieval: Given a question from \breakdataset{}, one retrieves similar training examples from an index of the training set. The question and training examples (the prompt) are passed to an inference LM that decodes the output.}
    \label{fig:overview}
\end{figure}
An attractive property of in-context learning is that
it provides a single model for multiple language understanding tasks. However,  \citet{Liu2021WhatMG} showed that downstream performance can vary widely depending on the choice of in-context examples. This has sparked interest in \emph{prompt retrieval} (see Fig.~\ref{fig:overview}), where given a test instance, training examples are chosen for the prompt based on some similarity metric. Recent work has either used off-the-shelf unsupervised similarity metrics, or trained a \emph{prompt retriever} to select examples based on surface similarity \cite{das-etal-2021-case}.

\begin{figure*}
    \centering
    \includegraphics[width=.95\textwidth]{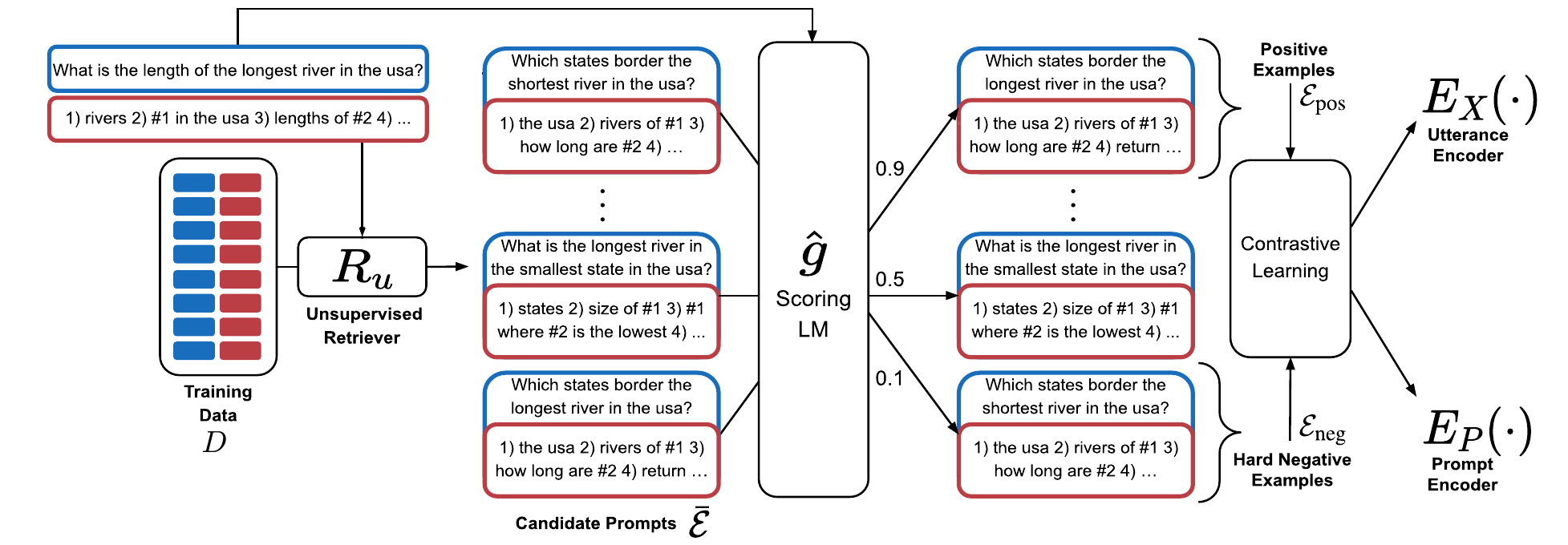}
    \caption{An overview of our approach for training EPR. Given a training example, we use an unsupervised retriever $R_u$ to obtain a set of candidates. We then pass the candidates to a scoring LM and label the top-$k$ and the bottom-$k$ as positive and negative examples, respectively. Last, we use this training data to train a dense retriever.}
    \label{fig:training}
\end{figure*}

In this work, we suggest to use \emph{language models themselves} to label examples that can serve as good prompts, and train a prompt retriever from this signal.
To train the retriever (see Fig.~\ref{fig:training}), we assume access to a training set of input-output pairs and to a \emph{scoring LM}, i.e., a language model that will be used to score prompts.
For each training example $(x,y)$, we go over other candidate training examples, and estimate the probability, according to the scoring LM, of $y$ conditioned on $x$ \emph{and} the candidate prompt.
We label training examples that lead to high probability as positive and low probability as negative and train a prompt retriever from this data using contrastive learning.
We argue that using an LM for labeling examples is a better proxy for training a retriever compared to previously-proposed surface similarity heuristics. Importantly, when creating the training data, we have access to the gold label $y$, which can be used to obtain a high-quality set of candidate prompts. This leads to good positive examples and hard negative examples, which are beneficial for training with a contrastive objective.



Using a scoring LM to train an efficient retriever for a potentially different test time \emph{inference LM} is beneficial in two scenarios. First, when the scoring LM is smaller than the inference LM and serves as a proxy for it. This results in cheap and efficient data generation for the retriever, accessible to a wide range of researchers. Second, our approach can be used even when the scoring and inference LMs are identical (e.g., both are GPT-3). This is beneficial when we do not have access to model parameters and can only use it as a service, an increasingly popular paradigm. In this case, we use the LM to
train a light-weight retriever that is only tasked with learning a similarity function.
More generally, given that the scale of LMs is likely to keep increasing in the foreseeable future, one can view our approach for \textbf{E}fficient \textbf{P}rompt \textbf{R}etrieval, or \emph{\ourmethod}, as a method for interfacing and learning to interact with large LMs.

We empirically test \ourmethod{} on three structured sequence-to-sequence tasks, where input natural language utterances are mapped to a meaning representation:
\mtop{} \citep{li-etal-2021-mtop} and \smcalflow \citep{andreas-etal-2020-task}, which focus on task-oriented dialogue, and \breakdataset{} \citep{wolfson-etal-2020-break}, a benchmark for mapping questions to a language-based meaning representation.
We observe that EPR substantially improves performance compared to prior work on prompt retrieval.
When the scoring LM and inference LM are identical (using \textsc{GPT-Neo} \citep{gpt-neo}), performance compared to the best baseline improves from 26\% to 31.9\% on \breakdataset{}, from 57\% to 64.2\% on \mtop{}, and from 51.4\% to 54.3\% on \smcalflow{}. When using \textsc{GPT-Neo} as a proxy for larger LMs (GPT-J, GPT-3, and \textsc{Codex}), we observe similar gains, where performance improves substantially in all cases. 


To conclude, we propose an approach for retrieving training examples for in-context learning in large language models, and show it substantially outperforms prior methods. Given recent developments in scaling LMs, designing efficient methods for interacting with LMs is an important direction for future research. 
Our code and data are publicly available at \url{https://github.com/OhadRubin/EPR}.


\section{Background: Prompt Retrieval}

\paragraph{Problem setup}
Given a training set $\sD = \{(x_i,y_i)\}_{i=1}^n$ of input-output sequences, and a test example  $x_\text{test}$, 
our goal is to train a retriever, $R(x_{\text{test}}, \sD)$, that will retrieve a subset of training examples 
$\sP = \{(x_j, y_j)\}_{j=1}^m \subset \mathcal \sD$, where $m \ll n$. We succinctly refer to $\sP$ as the \emph{prompt}.\footnote{\emph{Prompt} often refers to a natural language template filled by an input example \citep{liu2021pretrain}, but here it denotes the sequence of training examples provided as input to the LM.}

Given an inference LM, $g$, a good prompt
should lead to the target output sequence when the test example $x_\text{test}$ is concatenated to the prompt $\sP$ and passed as a prefix to $g$. Specifically, decoding from the LM $g([\sP; x_\text{test}])$ should yield $y_\text{test}$. In this work, we focus on structured tasks, such as semantic parsing, where $x$ is a natural language utterance and $y$ is a meaning representation for that utterance.

\paragraph{Prior work}
\newcite{Liu2021WhatMG} investigated the effect of different prompts on the performance of GPT-3 and demonstrated that the choice of in-context examples strongly affects downstream performance. They used an unsupervised sentence encoder to encode training examples, and retrieved for every test instance its nearest neighbors.

\newcite{das-etal-2021-case} trained a supervised prompt retriever for knowledge-base question answering. The retriever was trained with supervision that is tailored for knowledge-base queries, and relies on surface similarity between formal queries. Conversely, our approach takes advantage of the generative LM itself and is thus more general.



\newcite{shin-etal-2021-constrained} used GPT-3 to select examples for the prompt for few-shot semantic parsing. However, rather than training a retriever, they randomly sample a large set of utterance-program pairs from the training set, and choose those that are similar to the target instance question according to GPT-3. This results in an expensive inference procedure, where GPT-3 is run \textit{hundreds} of times for \textit{each} test instance, unlike our approach, which is based on a light-weight sub-linear retriever.



\section{Efficient Prompt Retriever}
\label{sec:method}

We now describe our method for training EPR, an efficient prompt retriever for in-context learning. We first describe how to generate labeled data (\S\ref{subsec:datageneration}), and then how to use the training data for training and inference (\S\ref{subsec:training}). Fig.~\ref{fig:training} provides an overview of the training procedure.
\subsection{Generating the Training Data}
\label{subsec:datageneration}

Our approach relies on finding which training examples can serve as good prompts for other training examples. Scoring all pairs of training examples is quadratic in $|\mathcal{D}|$, and thus prohibitive. Hence, we present a method for choosing 
a set of candidate examples $\bar{\sE} \subset D$, from which we will choose positive and negative examples for training.
Importantly, since we are \emph{not} at test time and are only generating data for training, 
we can use the target sequence $y$ to retrieve a good set of candidates. 
This can 
be approached using a simple retrieval method, given that our goal is to retrieve examples that are similar to the input in terms of their output sequence, $y$.

To obtain a high-quality candidate set of training examples, we take advantage of an unsupervised retriever, $\bar{\sE} = R_{u}((x,y),\mathcal{D})$. For the choice of the unsupervised retriever, we experiment with BM25 \cite{stephen2009bm25}, a sparse retriever that relies on surface text similarity, and SBERT \citep{reimers2019sentencebert}, which is based on dense sentence encoding. 
For both, we experimented with passing the retriever the training pair $(x,y)$ or the target sequence $y$ only, and found that using $y$ leads to slightly higher performance.


\paragraph{Scoring the candidate set}


Once we retrieve the set of candidates $\bar{\sE} = \{\bar{\eee}_1, \cdots, \bar{\eee_L}\}$ for a training example $(x,y)$,\footnote{We omit the dependence of $\bar{\sE}$ on $(x,y)$ for simplicity.} we score each candidate $\bar{\eee}_l \in \bar{\sE}$ independently with a scoring LM, $\hat{g}$, which serves as a proxy for the inference LM, $g$. Specifically, the score for a candidate prompt is
\begin{align*}
  s(\bar{\eee}_{l}) = \operatorname{Prob}_{\hat{g}}(y \mid \bar{\eee}_l, x),  
\end{align*}
which is the probability under the LM, $\hat{g}$, of the output sequence conditioned on the candidate prompt and input sequence. This indicates how helpful this candidate is for decoding the target (independent of all other candidates). We argue this score is a better proxy for the utility of a training example at inference time compared to prior approaches.

We apply this scoring function to all training examples, and define for each training example a set of positive examples $\sE_\text{pos}$, which includes the top-$k$ candidates in $\bar{\sE}$ according to $s(\bar{\eee}_l)$, and a set of negative examples  $\sE_\text{neg}$, which includes the bottom-$k$ candidates in $\bar{\sE}$ according to $s(\bar{\eee}_l)$. This should lead to relevant positive examples, assuming that the set of candidates, $\bar{\sE}$ includes good prompt candidates and hard negatives, since all candidates have high similarity with $(x,y)$ according to $R_u(y, \mathcal{D})$. 
With positive and negative examples at our disposal, we can now apply contrastive learning, which we describe next.

\subsection{Training and Inference}
\label{subsec:training}


\paragraph{Training}
Our training procedure proceeds exactly like the contrastive learning procedure from DPR \cite{karpukhin-etal-2020-dense}. This
procedure results in an input encoder $E_X(\cdot)$, which receives the sequence of input tokens, $x$, and a prompt encoder $E_P(\cdot)$, which receives a candidate prompt, namely, a concatenation of the tokens in an input-output pair. Both encoders are initialized with BERT-base \cite{devlin-etal-2019-bert}, and the output vector representation is given by the \texttt{CLS} token, as usual. The goal of training is to learn a similarity metric such that given a test example $x_\text{test}$, it will be similar to training examples that lead to decoding of $y_\text{test}$.

Our training instances are of the form $\langle x_i, \eee^+_{i}, \eee^-_{i,1},\dots \eee^-_{i,2B-1}\rangle$. Where the positive example $\eee^+_{i}$ is sampled from the set $\sE_{\text{pos}}^{(i)}$, and our negative examples consist of one hard negative example sampled from $\sE_\text{neg}^{(i)}$, $B-1$ positive examples from the other instances in the same mini-batch, and the $B-1$ hard negatives from those instances.
We define the similarity score between an input and an input-output pair to be the inner product $\mathrm{sim}(x, \eee)  =  E_X(x)^{\top}E_P(\eee)$. We can now define the typical contrastive learning objective and minimize for each example the negative log likelihood of the positive example:
\begin{eqnarray}
&& L(x_i, \eee^+_{i}, \eee^-_{i,1},\dots \eee^-_{i,2B-1}) \label{eq:training} \\
&=& -\log \frac{ e^{\mathrm{sim}(x_i, \eee_i^+)} } {e^{\mathrm{sim}(x_{i}, \eee_{i}^+)}+\sum_{j=1}^{2B-1}{e^{\mathrm{sim}(x_{i}, \eee_{i,j}^-)} }  }. \nonumber
\end{eqnarray}
An advantage of this approach is that for batch size $B$ the effective batch size is of order $B^2$, with the in-batch negatives trick \citep{henderson2017efficient}.

\paragraph{Inference } 
After training the input encoder and prompt encoder, we encode the entire set of training examples with $E_P(\cdot)$ in a pre-processing step using FAISS \citep{johnson2017billion}. 
At test time, given an input sequence, $x_\text{test}$, we compute its encoding $E_X(x_{\text{test}})$, and then use maximum inner-product search over the training data to find the $L$ most similar training examples, sorted by their inner product (from high to low): $\sP = (\eee_1, \dots, \eee_L)$. The final prompt $\sP'$ is determined by $C$, the maximal context size supported by 
the inference LM, $g$. Specifically, $L' \leq L$ is the largest $L'$ such $ \sum_{i=1}^{L'}|\eee_i| + |x_{\text{test}}| + |y'| \leq C$, where $|y'|$ is the desired maximal length of the generated output. Finally, we return the output of greedy decoding on $g([\eee_{L'}; \eee_{L'-1}; \dots; \eee_1; x_\text{test}])$.


We note that while at training time we score each training example independently, at test time the language model observes a prompt, i.e., \emph{a sequence} of examples. We leave modeling the dependence between different training examples to future work.

\section{Experimental Results}
\label{sec:experiments}

We now compare EPR to a wide range of unsupervised and supervised baselines, both when the scoring LM, $\hat{g}$, is smaller than the inference LM, $g$, and when they are identical.

\newcommand{\breakquestionA}[0]{\makecell[tl]{\emph{There are more birds in the image on}\\ \emph{the  right than in the image on the left.}}}
\newcommand{\breakmeaningA}[0]{\makecell[tl]{\texttt{1) return right image;}\\ \texttt{2) return birds in \#1;}\\ \texttt{3) return number of \#2;}\\ \texttt{4) return left image;}\\ \texttt{5) return birds in \#4}\\ \texttt{6) return number of \#5;}\\ \texttt{7) return if \#3 is higher than \#6;}}}
\newcommand{\mtopmeaningA}[0]{\makecell[tl]{\texttt{{[}IN:CREATE\_CALL =}\\ \texttt{\quad{[}SL:CONTACT = {[}IN:GET\_CONTACT =}\\ \texttt{\qquad{[}SL:CONTACT\_RELATED = Zoey{]}}\\ \texttt{\quad\qquad{[}SL:TYPE\_RELATION = wife{]}{]}{]}{]}}
}}
\newcommand{\mtopmeaningB}[0]{\makecell[tl]{
                             \texttt{{[}IN:GETWEATHER}\\ \texttt{\quad{[}SL:DATE\_TIME for March 13th {]} {]}}
}}
\newcommand{\smcalflowmeaningA}[0]{\makecell[tl]{
\texttt{(Yield (CreateCommitEventWrapper}\\ \texttt{\quad(CreatePreflightEventWrapper}\\ \texttt{\qquad(Event.start\_?}\\ \texttt{\quad\qquad(DateTimeConstraint (Morning)}\\ \texttt{\qquad\qquad(NextDOW (Thursday)))))))}
                            }                                }
\newcommand{\smcalflowmeaningB}[0]{\makecell[lt]{\texttt{(Yield(CreateCommitEventWrapper}\\ \texttt{\quad(CreatePreflightEventWrapper}\\ \texttt{\qquad(\& (Event.subject\_?}\\ \texttt{\quad\qquad(?= "lunch")) (Event.start\_?}\\ \texttt{\qquad\qquad(DateTimeConstraint}\\ \texttt{\quad\qquad\qquad(LateAfternoon) (Today)))))))}
}}

\newcommand{\smcalflowquestionA}[0]{\makecell[tl]{\emph{Can you create me a new meeting}\\\emph{on thursday morning?}}}
\begin{table*}[ht]
\centering
\footnotesize
\scalebox{.85}{
\begin{tabular}{@{}c|l|l|l@{}}
\toprule
\textbf{Dataset} & \makecell{\textbf{Size}} & \textbf{Utterance} & \textbf{Meaning Representation }                                  \\
\midrule
\breakdataset{} & 60K       &  \breakquestionA                                  & \breakmeaningA \\ 
\midrule
\multirow{1}{*}{\mtop{}} & \multirow{2}{*}{22K}        & \emph{call Zoey’s wife.}                                  & \mtopmeaningA \\
\midrule
\multirow{1}{*}{\smcalflow }  & \multirow{2}{*}{148K} & \smcalflowquestionA &                                       \smcalflowmeaningA \\
\bottomrule
\end{tabular}
}
\caption{Examples from each of the datasets we evaluate on.}
\label{tab:dataset-examples}
\end{table*}

\subsection{Datasets}

We focus on tasks that map utterances to meaning representations, where in-context examples can be used to learn the mapping from inputs to outputs. Examples from each dataset and the number of examples are in Table~\ref{tab:dataset-examples}. 
\begin{itemize}[leftmargin=*,itemsep=0pt,topsep=0pt]
 \item \textsc{Break} \citep{wolfson-etal-2020-break}:  A
 dataset mapping complex natural language questions into a language-based meaning representation, where a question is decomposed into an ordered list of atomic steps. We use the low-level \breakdataset{} subset, containing 44K/7K/8K examples in its training/development/test sets.
 
\item \mtop{} \citep{li-etal-2021-mtop}: A 
semantic parsing dataset, focused on task-oriented dialogue, where commands are mapped to complex nested queries across 11 domains. Similar to past work \cite{pasupat-etal-2021-controllable}, we use the English subset of \mtop{}, containing 16K/2K/4K examples in its training/development/test sets.

\item \smcalflow{} \citep{andreas-etal-2020-task}: A
large English-language task-oriented dataset that covers tasks such as calendar, weather, places, and people. The meaning representation is a dataflow program, which includes API calls, function composition and complex constraints. \smcalflow{} includes 15K development set examples and 134K training examples, from which we sample a random set of 44K examples for training.
\end{itemize}

\subsection{Baselines and Oracles}

We consider the following unsupervised baselines, which are applied at test time only.
\begin{itemize}[leftmargin=*,itemsep=0pt,topsep=0pt]

\item \textsc{Random}:
we randomly sample examples from the training set $\sD$.

\item \textsc{SBERT}:
We use \texttt{SentenceTransformers}, a library providing BERT-based sentence embeddings.\footnote{\url{https://www.sbert.net/index.html}.} Specifically, we use \texttt{paraphrase-mpnet-base-v2}, a 110M parameter model to encode the test utterance $x_\text{test}$ and retrieve the examples with the most similar utterances as in-context examples.

\item \textsc{BM25}:
We use the classical sparse retrieval method BM25
\cite{stephen2009bm25}, which is an extension of TF-IDF, to retrieve for each test utterance $x_\text{test}$ the training examples with the most similar utterance.

\item \textsc{BruteForce}:
We apply the prompt selection method for few-shot semantic parsing from \newcite{shin-etal-2021-constrained}. Given a test example $x_\text{test}$, we sample 200 training examples. For each training example $(x_i, y_i)$, compute $\operatorname{Prob}_{g}(x_{\text{test}} \mid x_i)$, and use the highest scoring examples for the prompt. Similar to us, this approach uses the inference LM to choose prompts. However, it does so at test time, which results in slow inference.
\end{itemize}

Next, we describe baselines that use the training set, $\sD$, to train a prompt retriever. All supervised methods share the following template. First, a candidate set $\bar{\sE}$ of $L=50$ examples is retrieved with the unsupervised retriever $R_u(y, \sD)$. We use BM25 as an unsupervised retriever, since it outperformed SBERT (see \S\ref{subsec:results}). 
We then score each candidate prompt $\bar{e}_l \in \bar{\sE}$ with \emph{some scoring function}, and label the top-$k$ prompts as positive examples and the bottom-$k$ as negative examples ($k=5$). Different supervised methods only differ in the scoring function itself.\footnote{Results for $k \in \{1, 5, 10\}$ and $L \in \{50, 100\}$ are in Appendix~\ref{sec:appendix}.}

\begin{itemize}[leftmargin=*,itemsep=0pt,topsep=0pt]
\item \textsc{DR-BM25}:
Here, we use the original BM25 scores for labeling positive and negative examples and training a dense retriever.

\item\textsc{Case-based Reasoning (CBR)}:
We adapt the scoring function from 
\newcite{das-etal-2021-case}, which focused on knowledge-base question answering. They define the weight for a pair of logical forms to be the F$_1$ score between the two sets of relations appearing in those logical forms, and use this weight to softly label their data.
Since in our setting we do not assume logical forms, we define the score between two output sequence $y_i$ and $y_j$ to be the F$_1$ between the two sets of tokens in $y_i$ and $y_j$, omitting stop words. 

\item \textsc{Efficient Prompt Retrieval (EPR)}:
Our full approach from \S\ref{sec:method}.
\end{itemize}

Last, we also consider two oracle models.

\begin{itemize}[leftmargin=*,itemsep=0pt,topsep=0pt]
\item \textsc{BM25-Oracle}:
We score test examples with BM25 using the \emph{gold} output sequence $R_{\text{BM25}}(y_\text{test}, \sD)$.
This provides an upper-bound on what can be learned by DR-BM25. EPR can potentially outperform this oracle, since its training signal goes beyond surface text similarity.

\item \textsc{LM-Oracle}:
We use the procedure for labeling training data at test time.
Given a test example $(x_\text{test}, y_\text{test})$, we first retrieve $L$ candidate training examples with $R_\text{BM25}(y_\text{test}, \sD)$, we then sort the candidates with the scoring LM $\hat{g}$, estimating the probability of $y_\text{test}$ given $x_\text{test}$ and the candidate prompt. This provides an upper bound for EPR, since EPR is trained to emulate this behaviour.
\end{itemize}

\subsection{Experimental Details}

\paragraph{Language models}
In this work, we only train a dense retriever, but use scoring and inference LMs. For our scoring LM, $\hat{g}$, we use \textsc{GPT-Neo} \citep{gpt-neo}, a 2.7B-parameter LM trained on The Pile \citep{gao2020pile}, an 825 GB English text corpus, constructed from a wide range of high-quality resources. 

In addition, we consider the following inference LMs:
(a) \textsc{GPT-J} \cite{gpt-j}: a 6B-parameter LM, also trained on The Pile. The advantage in this setup, is that \textsc{GPT-J} was trained on the same corpus as \textsc{GPT-Neo}.
(b) \textsc{GPT-3} \cite{brown2020language}: A 175B-parameter model, trained mostly on a filtered subset of common crawl.
(c) \textsc{Codex} \cite{Chen2021EvaluatingLL}: A GPT-3 175B-parameter model finedtuned on code from GitHub. Since our tasks involve mapping from utterances to programs or meaning representations, \textsc{Codex} might potentially perform well at in-context learning.

For all LMs, we use a maximum context size of $C=$2,048 tokens.

\paragraph{Evaluation}
On \breakdataset{}, we evaluate performance on the development set with LF-EM \citep{Hasson2021QuestionDW}, which is a better metric compared to Normalized Exact Match (NEM), the official metric, as it measures whether two meaning representations are semantically equivalent. On the test set, we use NEM.
On \mtop{} and \smcalflow{}, we evaluate with Exact Match (EM), i.e., whether the string output by the inference LM is identical to the reference string.

We evaluate EPR in two settings: (a) LM-as-a-service, and (b) LM-as-a-proxy. In the first setting, we use \textsc{GPT-Neo} as both the scoring LM and inference LM. In this setting, we evaluate on the full development sets of \breakdataset{}, \mtop{}, and \smcalflow{}. In the latter setting, as we access GPT-3 and \textsc{Codex} through a paid API, we sample a random subset of 1,000 development examples from each dataset and evaluate each model once on this subset.

\subsection{Results}
\label{subsec:results}

\begin{table}[t]
\centering
\scalebox{0.85}{
{\footnotesize
\begin{tabular}{llllc}
\toprule
& \textbf{Model}                                               & \textbf{\breakdataset{}}   & \textbf{\mtop{}}    & \textbf{\smcalflow{}} \\
\toprule
\multirow{3}{*}{\textbf{\textit{Unsuper.}}} & \textsc{Random}                                              & 1.7   & 7.3      & 8.9        \\
&\textsc{SBERT}                                     & 21.6  & 48.7  & 43.6    \\
&\textsc{BM25}                                     & 26.0     & 52.9  & 46.1    \\
&\textsc{Bruteforce}           & 7.7   & 18.1  & 11.1    \\
\midrule
\multirow{3}{*}{\textbf{\textit{Super.}}} & \textsc{DR-BM25}                                  & 23.6  & 50.2  & 43.1    \\
&\textsc{CBR}             & 25.7      & 57.0       &     51.4    \\
&\textsc{EPR} (ours) & \textbf{31.9 }    & \textbf{64.2 } & \textbf{54.3  }  \\
\midrule
\multirow{2}{*}{\textbf{\textit{Oracle}}} &\textsc{BM25-Oracle}                               & 32.3  & 58.9  & 47.3    \\
&\textsc{LM-Oracle}                   & 43.1  & 71.6  & 73.7    \\
\bottomrule
\end{tabular}
}}
\caption{Development results when \textsc{GPT-Neo} is the scoring and inference LM. Numbers for \breakdataset{} are LF-EM, and for \mtop{} and \smcalflow{} are EM.   }
\label{tab:results}
\end{table}

\begin{table}[t]
\centering
\scalebox{0.85}{
{\footnotesize
\begin{tabular}{llllc}
\toprule
& \textbf{Model}                                               & \textbf{\breakdataset{}}   & \textbf{\mtop{}} \\
\toprule
\multirow{1}{*}{\textbf{\textit{Unsuper.}}} & \textsc{BM25}                                              & 17.6   & 49.0              \\
\midrule
\multirow{2}{*}{\textbf{\textit{Super.}}} & \textsc{CBR}                                  & 18.4  & 57.5 \\
&\textsc{EPR} (ours) & \textbf{23.9}    & \textbf{64.4}   \\
\bottomrule
\end{tabular}
}}
\caption{Test results where \textsc{GPT-Neo} is the scoring and inference LM. Numbers for \breakdataset{} are NEM, the official metric, and for \mtop{} are EM.   }
\label{tab:testresults}
\end{table}

\paragraph{LM-as-a-service}
Table~\ref{tab:results} reports results where the scoring and inference LMs are identical. EPR substantially outperforms all other baselines. Specifically, when comparing to the best baseline, it improves performance from 26.0 to 31.9 on \breakdataset{}, from 57.0 to 64.2 on \mtop{}, and from 51.4 to 54.3 on \smcalflow{}. This shows that using the LM itself to label examples is an effective approach for obtaining a strong prompt retriever. 
Table~\ref{tab:testresults} shows test results on \breakdataset{} and \mtop{} corroborating that EPR substantially improves performance compared to BM25 and CBR.

For the unsupervised methods, 
the \textsc{Random} baseline shows that random sampling of training examples leads to poor performance. \textsc{BM25} outperforms \textsc{SBERT} for prompt retrieval, and consequently we use BM25 in all of our supervised approaches to retrieve the set of candidates, $\bar{\sE}$. Last, \textsc{BruteForce} performs worse than BM25. 
We assume this is since the training sets are large ($\sim$14-120K examples), and sampling 200 examples does not cover examples that are useful for \textsc{GPT-Neo}.

Interestingly, EPR outperforms \textsc{BM25-Oracle} on \mtop{} and \smcalflow{} and is comparable on \breakdataset{}. This is surprising since \textsc{BM25-Oracle} has access to the output sequence $y_\text{test}$ at test time, illustrating that the signal provided by the scoring LM for training goes beyond surface text similarity.
The performance of \textsc{LM-Oracle} is substantially higher than EPR, showing that
the supervision provided by the scoring LM is strong, and training a better retriever from this signal can substantially enhance performance.

\begin{table}[t]
\centering
\scalebox{0.9}{\footnotesize
\begin{tabular}{llll}
\toprule
                          & \textbf{Model}                                                            & \textbf{One-shot}                           & \textbf{Full-context} \\ \toprule
\multirow{2}{*}{\textbf{\textit{Unsuper.}}} & \multicolumn{1}{l|}{\textsc{Random}}            & \multicolumn{1}{l|}{1.1}           & 1.7          \\
                          & \multicolumn{1}{l|}{\textsc{BM25}}              & \multicolumn{1}{l|}{15.2}          & 26.0         \\ \hline 
\multirow{3}{*}{\textit{\textbf{Super.}}}   & \multicolumn{1}{l|}{\textsc{DR-BM25}}           & \multicolumn{1}{l|}{14.1}          & 23.6         \\
                          & \multicolumn{1}{l|}{\textsc{CBR}}               & \multicolumn{1}{l|}{14.5}          & 25.7         \\
                          & \multicolumn{1}{l|}{\textsc{EPR}}               & \multicolumn{1}{l|}{\textbf{23.0}} & 31.9         \\ \hline 
\multirow{3}{*}{\textbf{\textit{Oracle}}}   & \multicolumn{1}{l|}{\textsc{BM25-Oracle}}       & \multicolumn{1}{l|}{18.0}          & 32.3         \\
                          & \multicolumn{1}{l|}{\textsc{LM-Oracle}}         & \multicolumn{1}{l|}{33.3}          & 43.1         \\
                          & \multicolumn{1}{l|}{\textsc{Anycorrect-Oracle}} & \multicolumn{1}{l|}{53.6}          & -            \\ \bottomrule
\end{tabular}
}
\caption{Development results on \textsc{BREAK} with \textsc{GPT-Neo} in the one-shot setting. Numbers are \textsc{LF-EM}. Full-context is the corresponding numbers from Table~\ref{tab:results}. }

\label{tab:1prompts}

\end{table}

\begin{table*}[t]
\centering
\scalebox{0.9}{\footnotesize
\begin{tabular}{c|cccc|cccc|cccc}
\toprule
      \multicolumn{1}{c}{} & \multicolumn{4}{c}{\textbf{\breakdataset{}}} & \multicolumn{4}{c}{\textbf{\mtop{}}} & \multicolumn{4}{c}{\textbf{\smcalflow{}}} \\ \toprule
Method  & \textsc{Random}  & \textsc{BM25}   & \textsc{CBR} & \textsc{EPR}    & \textsc{Random} & \textsc{BM25}   & \textsc{CBR} & \textsc{EPR}    & \textsc{Random}   & \textsc{BM25}     & \textsc{CBR} & \textsc{EPR}     \\ \toprule
\textsc{GPT-3}   & 4.2\   & 20.1\ & 21.3\ & 25.3\ & 7.6\  & 52.5\ & 54.8\ & 62.6\ & 5.8\    & 35.3\ & 41.6\   & 46.5\  \\ 
\textsc{Codex}   & 8.9\   & 24.5\ & 24.2\ & 29.5\ & 10.8\ & 60.6\ & 59.4\ & 66.1\ & 7.2\    & 45.1\ & 48.7\   & 50.3\  \\ 
\textsc{GPT-J}   & 3.3\   & 26.7\ & 26.7\ & 31.5\ & 8.8\ & 56.6\ & 58.0\ & 65.4\ & 10.6\    & 50.4\ & 50.9\  & 57.4\  \\ 
\hline
\textsc{GPT-Neo} & 1.0\   & 22.8\ & 25.8\ & 29.9\ & 7.6\  & 52.8\ & 55.4\ & 63.6\ & 8.0\    & 46.1\ & 50.1\   & 53.5\  \\ \bottomrule
\end{tabular}
}
\caption{Results on a random sample of 1,000 examples from the development set when using GPT-Neo as a scoring LM across different inference LMs and datasets.}

\label{tab:gpt3}
\end{table*}
\newcommand{\BreakGoldUtt}[0]{\makecell[tl]{\textit{Give the code of the airport with the}\\ \textit{least flights.}}}
\newcommand{\BreakGoldMean}[0]{\makecell[tl]{\texttt{1) airports}\\ \texttt{2) flights of  \#1}\\ \texttt{3) number of  \#2 for each  \#1}\\\texttt{4) \#1 where \#3 is  lowest}\\\texttt{5) code of  \#4}}}
\newcommand{\BreakCbrUttA}[0]{\makecell[tl]{\textit{What destination has the fewest number}\\ \textit{of flights?}}}
\newcommand{\BreakCbrMeanA}[0]{\makecell[tl]{\texttt{1) destinations}\\\texttt{2) flights of \#1}\\\texttt{3) number of \#2 for each \#1}\\\texttt{4) \#1 where \#3 is lowest}}}
\newcommand{\BreakCbrUttB}[0]{\makecell[tl]{\textit{Which destination has least number of}\\ \textit{flights?}}}
\newcommand{\BreakCbrMeanB}[0]{\makecell[tl]{\texttt{1) destinations}\\\texttt{2) flights to \#1}\\\texttt{3) number of \#2 for each \#1}\\\texttt{4) \#1 where \#3 is lowest}}}
\newcommand{\BreakCbrUttC}[0]{\makecell[tl]{\textit{What is the number of airports per}\\ \textit{country, ordered from most to least?}}}
\newcommand{\BreakCbrMeanC}[0]{\makecell[tl]{\texttt{1) countries}\\\texttt{2) airports in \#1}\\\texttt{3) number of \#2 for each \#1}\\\texttt{4) \#3 sorted by most to least}}}
\newcommand{\BreakEprUttA}[0]{\makecell[tl]{\textit{What is the code of the city with the}\\ \textit{most students?}}}
\newcommand{\BreakEprMeanA}[0]{\makecell[tl]{\texttt{1) cities}\\\texttt{2) students in \#1}\\\texttt{3) number of \#2 for each \#1}\\\texttt{4) \#1 where \#3 is highest}\\\texttt{5) code of \#4}}}
\newcommand{\BreakEprUttB}[0]{\makecell[tl]{\textit{Return the code of the city that has the}\\ \textit{most students.}}}
\newcommand{\BreakEprMeanB}[0]{\makecell[tl]{\texttt{1) cities}\\\texttt{2) students in \#1}\\\texttt{3) number of \#2 for each \#1}\\\texttt{4) \#1 where \#3 is highest}\\\texttt{5) code of \#4}}}
\newcommand{\BreakEprUttC}[0]{\makecell[tl]{\textit{Find the count and code of the job has}\\ \textit{most employees.}}}
\newcommand{\BreakEprMeanC}[0]{\makecell[tl]{\texttt{1) jobs}\\\texttt{2) employees of \#1}\\\texttt{3) number of \#2 for each \#1}\\\texttt{4) \#1 where \#3 is highest}\\\texttt{5) employees of \#4}\\\texttt{6) number of \#5}\\\texttt{7) code of \#4}\\\texttt{8) \#6 , \#7}}}

\begin{table*}[ht]
\centering
\footnotesize
\scalebox{0.85}{
\begin{tabular}{cc|cl}
\multicolumn{1}{l}{}                        & \multicolumn{1}{l|}{}                                            & \multicolumn{1}{c|}{EPR}      & \multicolumn{1}{c}{CBR} \\ \hline
\multicolumn{1}{c|}{\multirow{2}{*}{\begin{tabular}[c]{@{}c@{}}Test\\ Example\end{tabular}}}  & Utterance                                                        & \multicolumn{2}{c}{\BreakGoldUtt}                             \\ \cline{2-4} 
\multicolumn{1}{c|}{}                       & \begin{tabular}[c]{@{}c@{}}Meaning\\ Representation\end{tabular} & \multicolumn{2}{c}{\BreakGoldMean}                            \\ \hline
\multicolumn{1}{c|}{\multirow{2}{*}{Top-1}} & Utterance                                                        & \multicolumn{1}{c|}{\BreakEprUttA}  & \BreakCbrUttA                 \\ \cline{2-4} 
\multicolumn{1}{c|}{}                       & \begin{tabular}[c]{@{}c@{}}Meaning\\ Representation\end{tabular} & \multicolumn{1}{c|}{\BreakEprMeanA} & \BreakCbrMeanA                \\ \hline
\multicolumn{1}{c|}{\multirow{2}{*}{Top-2}} & Utterance                                                        & \multicolumn{1}{c|}{\BreakEprUttB}  & \BreakCbrUttB                 \\ \cline{2-4} 
\multicolumn{1}{c|}{}                       & \begin{tabular}[c]{@{}c@{}}Meaning\\ Representation\end{tabular} & \multicolumn{1}{c|}{\BreakEprMeanB} & \BreakCbrMeanB                \\ \hline
\multicolumn{1}{c|}{\multirow{2}{*}{Top-3}} & Utterance                                               & \multicolumn{1}{c|}{\BreakEprUttC}  & \BreakCbrUttC                 \\ \cline{2-4} 
\multicolumn{1}{c|}{}                       & \begin{tabular}[c]{@{}c@{}}Meaning\\ Representation\end{tabular} & \multicolumn{1}{c|}{\BreakEprMeanC} & \BreakCbrMeanC                \\ \hline
\end{tabular}
}
\caption{An example from \breakdataset{} development set where EPR is correct and CBR is incorrect along with the top-3 training examples retrieved from each retriever.}
\label{tab:retrieval-examples}
\end{table*}

We further evaluate our models in the one-shot setup, i.e., when the prompt given to the inference LM includes the highest scoring example only. In this setup, the inference LM is applied in the same setting as when we generate labeled data, where we go over each prompt candidate independently. Since train and test time are now closer, we can expect the advantage of EPR to be more pronounced.

Table~\ref{tab:1prompts} shows the results. Indeed, EPR outperforms the best baseline by 8.5\%, and \textsc{BM25-Oracle} by 5\%. In addition, we examine \textsc{AnyCorrect-Oracle}, which tests whether any of the candidates returned by BM25 leads to the correct output. \textsc{AnyCorrect-Oracle} reaches 53.6\%, 20 points above \textsc{LM-Oracle}. This shows the high quality of candidates provided by BM25 (applied on the $y$), as one can reach more than 50\% LF-EM with just a single prompt. Moreover, it hints that a better scoring function can potentially further improve performance.

\paragraph{LM-as-a-proxy}

Table~\ref{tab:gpt3} shows results when the scoring LM is \textsc{GPT-Neo} and the inference LM is a larger LM. First, the trends are similar to the LM-as-a-service setup, i.e., EPR substantially outperforms prior baselines, including our best unsupervised baseline, BM25, and the best supervised baseline, CBR, by 2-8 points on all datasets and all pre-trained models. Thus, \textsc{GPT-Neo} serves as a good proxy for choosing training examples.

To further validate this finding, we evaluate the performance of \textsc{GPT-J} on \breakdataset{} with \textsc{GPT-Neo} as the scoring LM compared to using \textsc{GPT-J} itself as the scoring LM. We find performance improves slightly from 31.5 to 33.6. 
Analogously, when using \textsc{Codex} as the scoring LM and inference LM performance remains roughly the same: 29.5$\rightarrow$29.3. Thus, using a smaller LM (\textsc{GPT-Neo}) is an effective strategy for training a retriever that will be applied on other LMs.
Zooming in on different inference LMs, \textsc{GPT-J} performs slightly better than \textsc{GPT-Neo} across the board, since it was trained on the same data and using the same procedure as \textsc{GPT-Neo}. \textsc{Codex} outperforms GPT-3, which can be explained by the fact that it was trained on code, and our datasets involve mapping to programs or meaning representations. Surprisingly, \textsc{GPT-J} outperforms \textsc{Codex} (except on \mtop{}) and GPT-3 despite being 30x smaller. This can perhaps be explained by the fact that GPT-J was trained on a different dataset (The Pile \citep{gao2020pile}).



\begin{table}[ht]
\centering
\footnotesize
\scalebox{0.83}{
\begin{tabular}{cl|cc|cc|l}
\toprule
             &  \multirow[c]{2}{*}{\textbf{Pattern}}  & \multicolumn{2}{c}{\textbf{Copied}} & \multicolumn{2}{c}{\textbf{Novel}} &  \textbf{Total} \\
             &  &    Acc &   Rate &    Acc &   Rate &    Acc \\
\midrule
 \multirow[c]{2}{*}{\breakdataset{}}  & Exact &  55.1\% &  10.4\% &  29.7\% &  89.6\% &  \multirow[c]{2}{*}{32.3\%} \\
             & Abstract &  58.0\% &  41.1\% &  14.5\% &  58.9\% &  {} \\
\midrule
 \multirow[c]{2}{*}{\mtop{}} & Exact &  77.3\% &  25.3\% &  59.7\% &  74.7\% &  \multirow[c]{2}{*}{64.2\%}  \\
             & Abstract &  71.6\% &  84.5\% &  23.4\% &  15.5\% &  {} \\
\midrule
\multirow[c]{2}{*}{\textsc{SmCal}}  & Exact &  62.5\% &  60.2\% &  42.4\% &  39.8\% &  \multirow[c]{2}{*}{54.5\%} \\
             & Abstract &  62.4\% &  81.2\% &  20.6\% &  18.8\% &  {} \\
\bottomrule
\end{tabular}
}
\caption{Accuracy comparison between the decoded instances that contained patterns from the prompt and novel instances those that don't. Results shown are on the LM-as-a-service setup using \textsc{GPT-Neo}. }
\label{tab:copy_acc_rate}
\end{table}



\begin{figure}
    \noindent
    \includegraphics[width=0.4\textwidth,center]{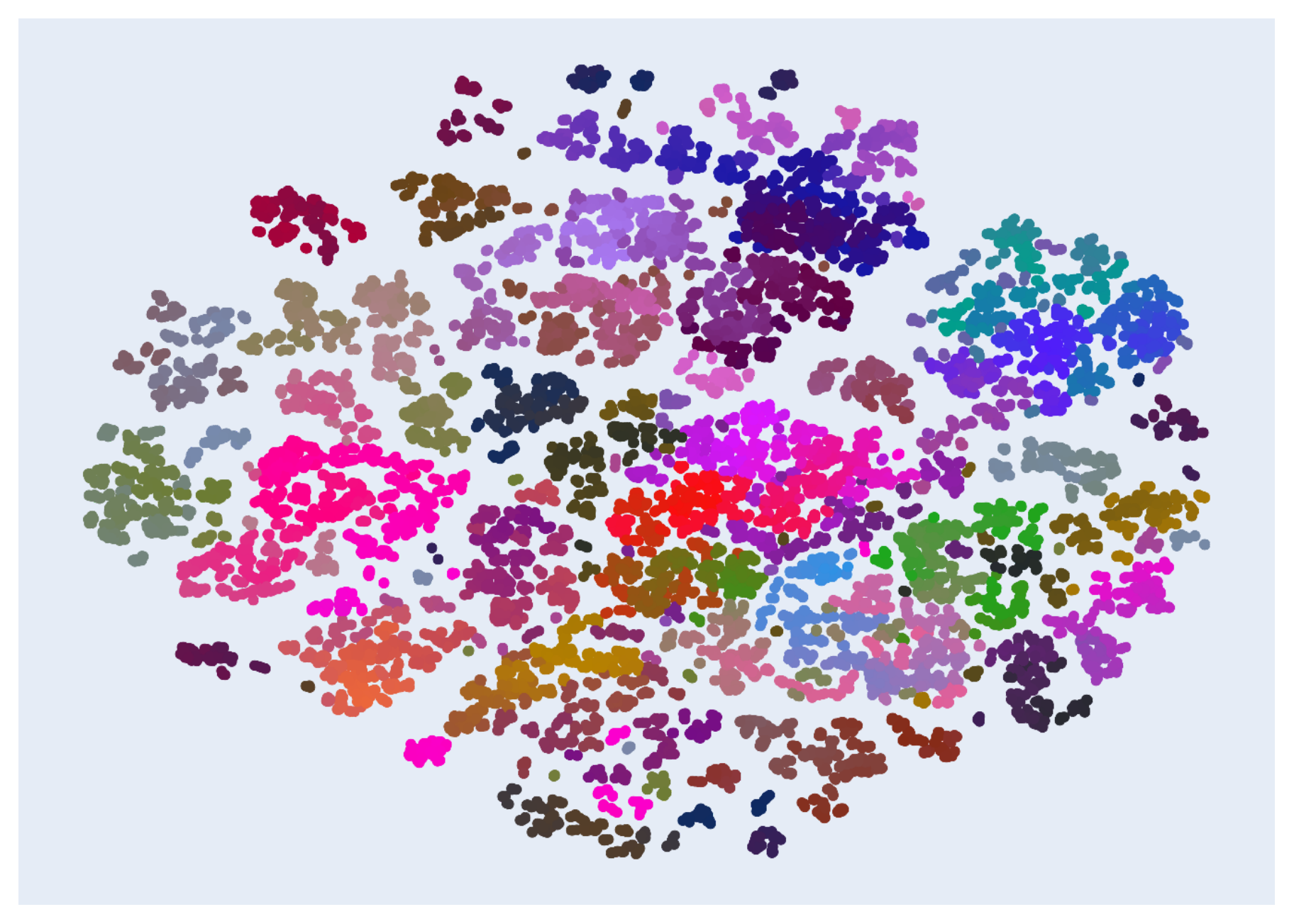}
    \caption{A t-SNE projection and clustering of the representations learned by EPR for the training examples in \breakdataset{}. An interactive version displaying individual examples is available \hyperblue{https://anonymous.4open.science/w/Learning-to-retrieve-prompts-for-in-context-learning-1C4F/}{here}. }
    \label{fig:tsne}
\end{figure}

\paragraph{Analysis}

Table~\ref{tab:retrieval-examples} shows an example from \breakdataset{} where EPR decodes the correct output, while CBR does not. All training examples retrieved by EPR perform an argmax (argmin in the original utterance), and return in the final step \emph{``a code''}, while the third example retrieved by CBR does not perform an argmax or argmin, and do not involve \emph{``a code''}.
We provide additional examples in App.~\ref{sec:appendix}.

Figure~\ref{fig:tsne} shows a t-SNE \citep{Hinton2002StochasticNE} projection of the embeddings learned by EPR for the training examples of \breakdataset{}, with a link to an interactive version, where we applied the OPTICS \cite{optics1992ankerst,Schubert2018ImprovingTC} clustering algorithm. Examining clusters shows that EPR captures both lexical and structure similarity. Examples for 
clusters are also available in App.~\ref{sec:appendix}.

\begin{figure}
    \noindent
    \includegraphics[width=0.45\textwidth]{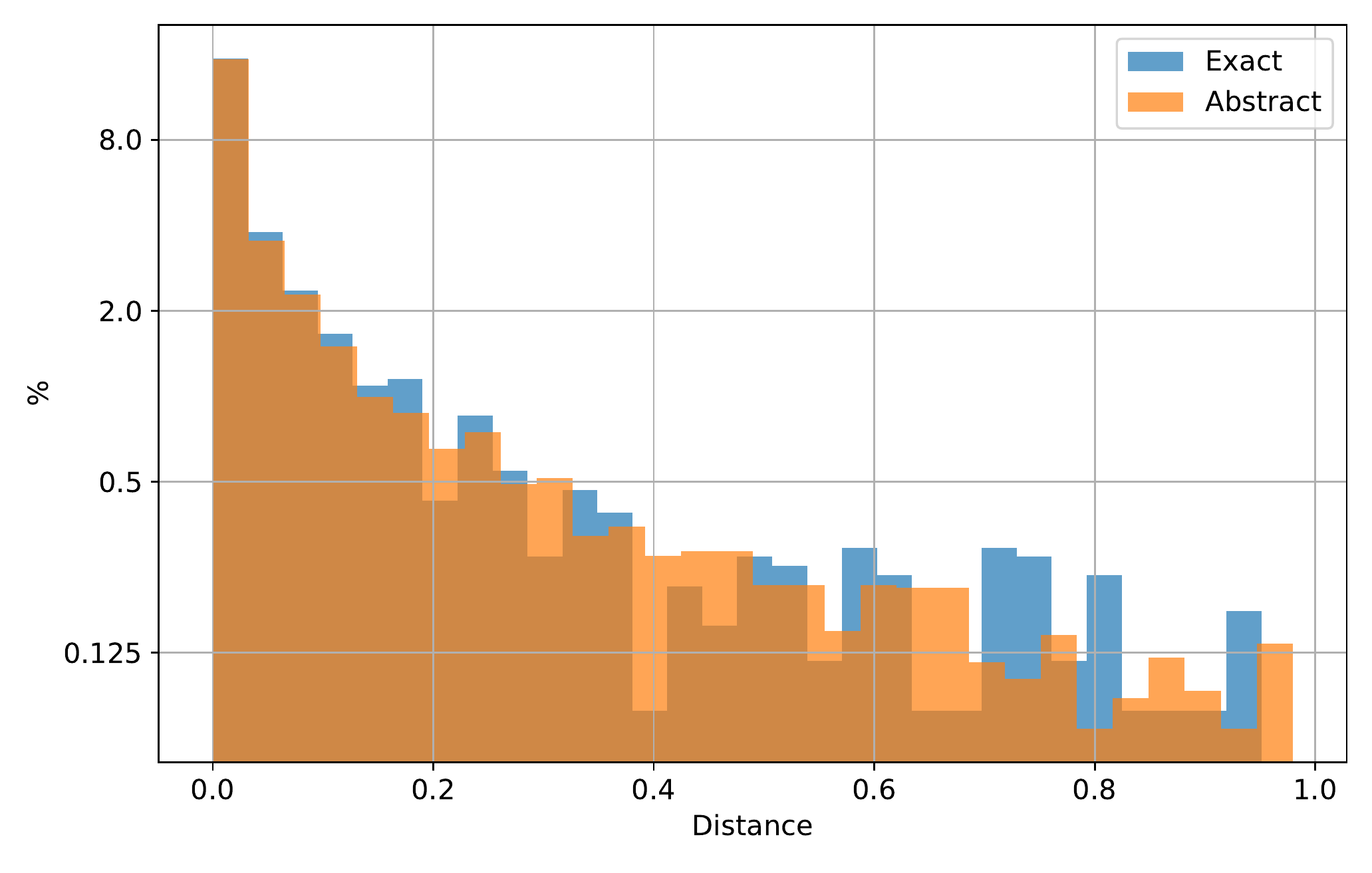}
    \caption{On the subset of copied patterns we plot the distribution of the distance from the test instance to the example containing the pattern. Shown on the \breakdataset{} validation set using EPR in the LM-as-a-service setup using \textsc{GPT-Neo}. Note that the y-axis is in log-scale.}
    \label{fig:hist_percentile}
\end{figure}


\paragraph{Prompt copying}
We analyze how the LM utilizes in-context prompts. Specifically, is the target output copied from one of the prompts or is it a composition of different prompt fragments, which result in generalization to new structures.

To achieve this, we define two types of copying. (a) \textit{Exact copying} measures if the generated output exactly matches one of the examples in the prompt, and (b) \textit{Abstract copying}, that quantifies if the structure of the decoded output matches any of the structures seen in the prompt. Specifically, we eliminate the effect of non-structural elements such as entities and function arguments. We replace every sequence of words in the logical form that appears in the input utterance with the string \emph{[MASKED]} for both the target utterance and in-context examples. If the masked logical form that the LM decoded appears in the set of masked examples defined by the prompt, we say that the LM copied that abstract pattern.

Table~\ref{tab:copy_acc_rate} presents the results on the validation set for each of our three datasets, as well as the accuracy on each subset.
We observe that the rate of copying is much higher in \mtop{} and \smcalflow{} compared to \breakdataset{}, where in \mtop{} and \smcalflow{} abstract copying reaches more than 80\%. Moreover, accuracy on examples where copying occurred is much higher compared to accuracy where no copying happened.
For example, on \mtop{}, 84.5\% of the examples were abstractly copied, and on that subset of examples, EPR achieves 71.6\% EM, compared to 64.2\% on the entire validation set. 
Nevertheless, even though accuracy is much lower in cases where no copying occurred, accuracy is not negligible, which shows that some form of generalization to new structures is taking place.

Another follow-up question is whether the model copies patterns from prompts uniformly or does it attend mostly to the ones with high retrieval score. To answer this, we look at  the subset of examples where copying occurred. We then identify for each example the highest-ranking prompt that was copied from, and define the \emph{distance} of that prompt by dividing the rank by the number of prompts that fit in that example. 
Figure~\ref{fig:hist_percentile} shows the distribution over distances for the \breakdataset{} dataset.
We observe that copying happens mostly from highly-ranked prompts.





\section{Related Work}
\paragraph{In-context learning}
Our understanding of in-context learning has grown substantially recently. \newcite{saunshi2021a} suggests that by conditioning on a prompt, the task of predicting the next word approaches linear separability. \newcite{Xie2021AnEO} suggests that in-context learning occurs when the model infers a shared latent concept between examples in a prompt. \newcite{Levine2021TheIB} present a pre-training scheme  theoretically motivated by the bias of in-context learning, that gives significant improvements. Recently, \newcite{Min2022RethinkingTR} showed that the model does not rely on the ground truth input-label mapping provided in the demonstrations as much as previously thought.

\paragraph{Retrieval}
Research on training dense retrievers has skyrocketed recently, propelled by interest in open-domain question answering \cite{chen-etal-2017-reading,lee-etal-2019-latent,karpukhin-etal-2020-dense,Guu2020RetrievalAL,khattab2020colbert,qu-etal-2021-rocketqa}.
Work on retrieval-based methods has also spread more widely to other knowledge-intensive tasks \cite{lewis2020rag}, e.g., fact verification \cite{samarinas-etal-2021-improving}.

Similar to us, \newcite{pasupat-etal-2021-controllable} proposed to use retrieval in semantic parsing. However, they focus on \emph{controlling} the output generated by a model. Retrieval methods have also been successfully used in language modeling \cite{khandelwal20generalization,borgeaud2021improving,Alon2022NeuroSymbolicLM} and machine translation \cite{khandelwal2021nearest}.

\paragraph{Prompts} Developing methods for interacting with LMs and extracting desired behaviours has attracted considerable attention, under the  umbrella term \emph{prompting}. In this work, prompts are a set of in-context training examples, but substantial effort has also been devoted to casting natural language tasks as language modeling by phrasing the target task in natural language (see survey in \cite{liu2021pretrain}). Such approaches include prompt engineering through manual patterns  \cite{petroni-etal-2019-language,schick-schutze-2021-exploiting}, decoding methods \citep{min2021noisy,zhao2021calibrate,holtzman-etal-2021-surface}, and methods for extracting either hard \cite{shin-etal-2020-autoprompt,haviv-etal-2021-bertese} or soft \cite{li-liang-2021-prefix,zhong-etal-2021-factual,qin-eisner-2021-learning} prompts automatically.

\paragraph{Prompt retrieval for supervised models}

In parallel to this work, adding training examples as additional input has been shown to be useful for supervised models as well. \newcite{Wang2022TrainingDI} and \newcite{xu2021human} used BM25 to retrieve and augment the input with similar examples from the training set. Fine-tuning the model with the additional inputs improved performance on tasks such as summarization and question answering. Such methods can also potentially benefit from a stronger retriever.

\section{Conclusions}

Large pre-trained LMs are becoming an inseparable part of the natural language understanding eco-system. However, accessing their weights or updating them can be prohibitive for many researchers. In this work, we propose EPR, a method for learning to retrieve good prompts for in-context learning, by using \emph{language models themselves} as the scoring function. This allows us to train a light-weight retriever and substantially improve performance on three challenging tasks. 

More broadly, given that large LMs models are likely to play a prominent role in developing language understanding models, it is important to develop approaches for interacting with such models effectively. EPR can be viewed as a step in this direction.

\section*{Acknowledgement}

We thank Ori Ram and Itay Itzhak for helpful suggestions and meaningful discussions. This research was supported in part by The Yandex Initiative for Machine Learning, and The European Research Council (ERC) under the European Union Horizons 2020 research and innovation programme (grant ERC DELPHI 802800). This work was completed in partial fulfillment for the Ph.D degree of Ohad Rubin.

\bibliography{custom}
\bibliographystyle{acl_natbib}

\appendix

\section{Appendix}
\label{sec:appendix}

\begin{figure*}
    \noindent

    \includegraphics[width=1.0\textwidth]{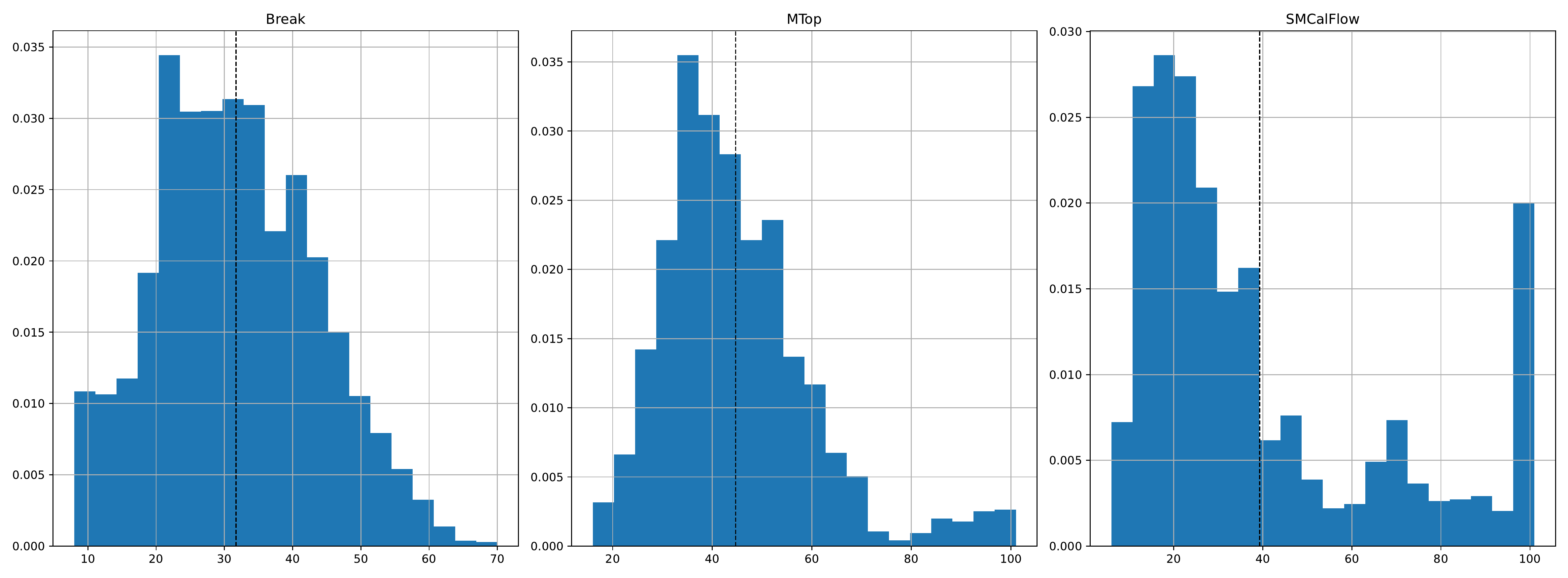}

    \caption{Distribution of the number of in-context examples per test instance for each of the datasets. We mark the distribution mean using a dashed line. }
    \label{fig:n_prompt_hist}
\end{figure*}
\paragraph{Distribution of the number of in-context examples} Since the selection procedure for in-context examples is dynamic, the number of in-context examples differs for each test instance.  In Figure~\ref{fig:n_prompt_hist}, we plot the histogram of the number of examples we fit in $C=2,048$ tokens.

\paragraph{Effect of hyperparameters}
We test the effect of $k$, the number of prompts labeled as positive or negative, and $L$, the number of prompts retrieved by the unsupervised retriever. Table~\ref{tab:diff_k} shows that performance is is generally robust w.r.t these hyperparameters.

\begin{table}[ht]
\centering
\footnotesize
\scalebox{1.1}{

\begin{tabular}{llll}
\toprule
  & \breakdataset{} & \mtop{} & \smcalflow{} \\
\midrule
 $k=1$ &            31.5\% &     63.0\% &        54.5\% \\
 $k=5$ &           31.9\% &   64.2\% &        54.3\% \\
$k=10$ &             31.0\% &   64.1\% &        52.2\% \\
\midrule
 $L=50$ &           31.9\% &   64.2\% &        54.3\% \\
$L=100$ &           32.3\% &   63.7\% &          51.0\% \\
\bottomrule
\end{tabular}
}
\caption{In the LM-as-a-service setup, using GPT-Neo, we search for other values for $L$ and $k$, and note that the choice of our hyperparameters is robust.}
\label{tab:diff_k}
\end{table}

\paragraph{Training details}
To train EPR, we use the Adam optimizer \cite{kingma2014adam} with batch size 120 and learning rate 1e-4 on eight RTX 3090. We run training for 30 epochs. We used the default DPR hyperparameters without tuning. We used the final epoch of the model to perform model selection, and applied minimal learning rate tuning on the validation set of \breakdataset{}.

\paragraph{Risk assessment}
Large language models have been shown to exhibit various kinds of bias \citep{stoca2021bender}, since EPR is trained on the signal obtained from such large LMs, it might also exhibit these biases.

\paragraph{Additional examples}
Tables~\ref{tab:mtop-retrieval-examples}, \ref{tab:Smcalflow-retrieval-examples}, and~\ref{tab:smcalflowB-retrieval-examples} provide more examples for cases where EPR is correct while CBR is incorrect along with the top-3 prompts for each method.

\newcommand{\MtopGoldUtt}[0]{\makecell[tl]{\textit{Remind me to add 2 dozen eggs to my}\\ \textit{grocery list.}}}
\newcommand{\MtopGoldMean}[0]{\makecell[tl]{\texttt{{[}IN:CREATE\_REMINDER {[}SL:PERSON\_REMINDED}\\ \texttt{me {]} {[}SL:TODO add 2 dozen eggs to my}\\ \texttt{grocery list {]} {]}}}}
\newcommand{\MtopCbrUttA}[0]{\makecell[tl]{\textit{Please add a grocery list to my list of}\\ \textit{things to be reminded about doing today.}}}
\newcommand{\MtopCbrMeanA}[0]{\makecell[tl]{\texttt{{[}IN:CREATE\_REMINDER {[}SL:TODO a grocery}\\ \texttt{list {]} {[}SL:PERSON\_REMINDED my {]}}\\ \texttt{{[}SL:DATE\_TIME today {]} {]}}}}
\newcommand{\MtopCbrUttB}[0]{\makecell[tl]{\textit{Remind me to make a grocery list}}}
\newcommand{\MtopCbrMeanB}[0]{\makecell[tl]{\texttt{{[}IN:CREATE\_REMINDER {[}SL:PERSON\_REMINDED}\\ \texttt{me {]} {[}SL:TODO make a grocery list {]} {]}}}}
\newcommand{\MtopCbrUttC}[0]{\makecell[tl]{\textit{I need to make a grocery list; will you}\\ \textit{remind me when I get off work at 5:00}\\ \textit{p.m.?}}}
\newcommand{\MtopCbrMeanC}[0]{\makecell[tl]{\texttt{{[}IN:CREATE\_REMINDER {[}SL:TODO make a}\\ \texttt{grocery list {]} {[}SL:PERSON\_REMINDED me {]}}\\ \texttt{{[}SL:DATE\_TIME at 5 : 00 p.m . {]} {]}}}}
\newcommand{\MtopEprUttA}[0]{\makecell[tl]{\textit{Remind me to get two bottles of water.}}}
\newcommand{\MtopEprMeanA}[0]{\makecell[tl]{\texttt{{[}IN:CREATE\_REMINDER {[}SL:PERSON\_REMINDED}\\ \texttt{me {]} {[}SL:TODO get two bottles of water {]}}\\ \texttt{{]}}}}
\newcommand{\MtopEprUttB}[0]{\makecell[tl]{\textit{Remind me to bring an extra pair of}\\ \textit{shoes to the river.}}}
\newcommand{\MtopEprMeanB}[0]{\makecell[tl]{\texttt{{[}IN:CREATE\_REMINDER {[}SL:PERSON\_REMINDED}\\ \texttt{me {]} {[}SL:TODO bring an extra pair of}\\ \texttt{shoes to the river {]} {]}}}}
\newcommand{\MtopEprUttC}[0]{\makecell[tl]{\textit{Remind me to add bottled water to}\\ \textit{grocery list.}}}
\newcommand{\MtopEprMeanC}[0]{\makecell[tl]{\texttt{{[}IN:CREATE\_REMINDER {[}SL:PERSON\_REMINDED}\\ \texttt{me {]} {[}SL:TODO add bottled water to}\\ \texttt{grocery list {]} {]}}}}

\begin{table*}[]
\centering
\footnotesize
\scalebox{0.9}{
\begin{tabular}{cc|cl}
\multicolumn{1}{l}{}                        & \multicolumn{1}{l|}{}                                            & \multicolumn{1}{c|}{EPR}      & \multicolumn{1}{c}{CBR} \\ \hline
\multicolumn{1}{c|}{\multirow{2}{*}{\begin{tabular}[c]{@{}c@{}}Test\\ Example\end{tabular}}}  & Utterance                                                        & \multicolumn{2}{c}{\MtopGoldUtt}                             \\ \cline{2-4} 
\multicolumn{1}{c|}{}                       & \begin{tabular}[c]{@{}c@{}}Meaning\\ Representation\end{tabular} & \multicolumn{2}{c}{\MtopGoldMean}                            \\ \hline
\multicolumn{1}{c|}{\multirow{2}{*}{Top-1}} & Utterance                                                        & \multicolumn{1}{c|}{\MtopEprUttA}  & \MtopCbrUttA                 \\ \cline{2-4} 
\multicolumn{1}{c|}{}                       & \begin{tabular}[c]{@{}c@{}}Meaning\\ Representation\end{tabular} & \multicolumn{1}{c|}{\MtopEprMeanA} & \MtopCbrMeanA                \\ \hline
\multicolumn{1}{c|}{\multirow{2}{*}{Top-2}} & Utterance                                                        & \multicolumn{1}{c|}{\MtopEprUttB}  & \MtopCbrUttB                 \\ \cline{2-4} 
\multicolumn{1}{c|}{}                       & \begin{tabular}[c]{@{}c@{}}Meaning\\ Representation\end{tabular} & \multicolumn{1}{c|}{\MtopEprMeanB} & \MtopCbrMeanB                \\ \hline
\multicolumn{1}{c|}{\multirow{2}{*}{Top-3}} & \textit{Utterance}                                               & \multicolumn{1}{c|}{\MtopEprUttC}  & \MtopCbrUttC                 \\ \cline{2-4} 
\multicolumn{1}{c|}{}                       & \begin{tabular}[c]{@{}c@{}}Meaning\\ Representation\end{tabular} & \multicolumn{1}{c|}{\MtopEprMeanC} & \MtopCbrMeanC                \\ \hline
\end{tabular}
}
\caption{An example from \mtop{} development set where EPR is correct and CBR is incorrect along with the top-3 training examples retrieved from each retriever.}
\label{tab:mtop-retrieval-examples}
\end{table*}

\newcommand{\SmcalflowGoldUtt}[0]{\makecell[tl]{\textit{confirmed thanks}}}
\newcommand{\SmcalflowGoldMean}[0]{\makecell[tl]{\texttt{(PleasantryAnythingElseCombined)}}}
\newcommand{\SmcalflowCbrUttA}[0]{\makecell[tl]{\textit{Yes, but make sure to let me know the}\\ \textit{weather for that time.}}}
\newcommand{\SmcalflowCbrMeanA}[0]{\makecell[tl]{\texttt{(let (x0 (Execute (\textasciicircum{}(Dynamic)}\\ \texttt{ConfirmAndReturnAction))) (do (Yield x0)}\\ \texttt{(Yield (WeatherForEvent (\textasciicircum{}(Dynamic) item}\\ \texttt{x0)))))}}}
\newcommand{\SmcalflowCbrUttB}[0]{\makecell[tl]{\textit{Awesome, perfect}}}
\newcommand{\SmcalflowCbrMeanB}[0]{\makecell[tl]{\texttt{(Yield (Execute (\textasciicircum{}(Dynamic)}\\ \texttt{ConfirmAndReturnAction)))}}}
\newcommand{\SmcalflowCbrUttC}[0]{\makecell[tl]{\textit{Perfect...}}}
\newcommand{\SmcalflowCbrMeanC}[0]{\makecell[tl]{\texttt{(Yield (Execute (\textasciicircum{}(Dynamic)}\\ \texttt{ConfirmAndReturnAction)))}}}
\newcommand{\SmcalflowEprUttA}[0]{\makecell[tl]{\textit{it's ok bye}}}
\newcommand{\SmcalflowEprMeanA}[0]{\makecell[tl]{\texttt{(PleasantryAnythingElseCombined)}}}
\newcommand{\SmcalflowEprUttB}[0]{\makecell[tl]{\textit{It's ok}}}
\newcommand{\SmcalflowEprMeanB}[0]{\makecell[tl]{\texttt{(PleasantryAnythingElseCombined)}}}
\newcommand{\SmcalflowEprUttC}[0]{\makecell[tl]{\textit{It's ok}}}
\newcommand{\SmcalflowEprMeanC}[0]{\makecell[tl]{\texttt{(PleasantryAnythingElseCombined)}}}

\begin{table*}[]
\centering
\footnotesize
\scalebox{0.9}{
\begin{tabular}{cc|cl}
\multicolumn{1}{l}{}                        & \multicolumn{1}{l|}{}                                            & \multicolumn{1}{c|}{EPR}      & \multicolumn{1}{c}{CBR} \\ \hline
\multicolumn{1}{c|}{\multirow{2}{*}{\begin{tabular}[c]{@{}c@{}}Test\\ Example\end{tabular}}}  & Utterance                                                        & \multicolumn{2}{c}{\SmcalflowGoldUtt}                             \\ \cline{2-4} 
\multicolumn{1}{c|}{}                       & \begin{tabular}[c]{@{}c@{}}Meaning\\ Representation\end{tabular} & \multicolumn{2}{c}{\SmcalflowGoldMean}                            \\ \hline
\multicolumn{1}{c|}{\multirow{2}{*}{Top-1}} & Utterance                                                        & \multicolumn{1}{c|}{\SmcalflowEprUttA}  & \SmcalflowCbrUttA                 \\ \cline{2-4} 
\multicolumn{1}{c|}{}                       & \begin{tabular}[c]{@{}c@{}}Meaning\\ Representation\end{tabular} & \multicolumn{1}{c|}{\SmcalflowEprMeanA} & \SmcalflowCbrMeanA                \\ \hline
\multicolumn{1}{c|}{\multirow{2}{*}{Top-2}} & Utterance                                                        & \multicolumn{1}{c|}{\SmcalflowEprUttB}  & \SmcalflowCbrUttB                 \\ \cline{2-4} 
\multicolumn{1}{c|}{}                       & \begin{tabular}[c]{@{}c@{}}Meaning\\ Representation\end{tabular} & \multicolumn{1}{c|}{\SmcalflowEprMeanB} & \SmcalflowCbrMeanB                \\ \hline
\multicolumn{1}{c|}{\multirow{2}{*}{Top-3}} & \textit{Utterance}                                               & \multicolumn{1}{c|}{\SmcalflowEprUttC}  & \SmcalflowCbrUttC                 \\ \cline{2-4} 
\multicolumn{1}{c|}{}                       & \begin{tabular}[c]{@{}c@{}}Meaning\\ Representation\end{tabular} & \multicolumn{1}{c|}{\SmcalflowEprMeanC} & \SmcalflowCbrMeanC                \\ \hline
\end{tabular}
}
\caption{An example from \smcalflow{} development set where EPR is correct and CBR is incorrect along with the top-3 training examples retrieved from each retriever.}
\label{tab:Smcalflow-retrieval-examples}
\end{table*}

\newcommand{\SmcalflowBGoldUtt}[0]{\makecell[tl]{\textit{Create a meeting with David Crim today}}}
\newcommand{\SmcalflowBGoldMean}[0]{\makecell[tl]{\texttt{(Yield (CreateCommitEventWrapper}\\ \texttt{(CreatePreflightEventWrapper (\&}\\ \texttt{(Event.start\_? (DateTime.date\_? (?=}\\ \texttt{(Today)))) (Event.attendees\_?}\\ \texttt{(AttendeeListHasRecipient (Execute}\\ \texttt{(refer (extensionConstraint}\\ \texttt{(RecipientWithNameLike (\textasciicircum{}(Recipient)}\\ \texttt{EmptyStructConstraint) (PersonName.apply}\\ \texttt{"David Crim")))))))))))}}}
\newcommand{\SmcalflowBCbrUttA}[0]{\makecell[tl]{\textit{set up a meeting with both of David}\\ \textit{Crim's reports today}}}
\newcommand{\SmcalflowBCbrMeanA}[0]{\makecell[tl]{\texttt{(Yield (CreateCommitEventWrapper}\\ \texttt{(CreatePreflightEventWrapper (\&}\\ \texttt{(Event.start\_? (DateTime.date\_? (?=}\\ \texttt{(Today)))) (Event.attendees\_?}\\ \texttt{(AttendeeListHasPeople (FindReports}\\ \texttt{(Execute (refer (extensionConstraint}\\ \texttt{(RecipientWithNameLike (\textasciicircum{}(Recipient)}\\ \texttt{EmptyStructConstraint) (PersonName.apply}\\ \texttt{"David Crim"))))))))))))}}}
\newcommand{\SmcalflowBCbrUttB}[0]{\makecell[tl]{\textit{Make a meeting with David Largenstop on}\\ \textit{the 24th.}}}
\newcommand{\SmcalflowBCbrMeanB}[0]{\makecell[tl]{\texttt{(Yield (CreateCommitEventWrapper}\\ \texttt{(CreatePreflightEventWrapper (\&}\\ \texttt{(Event.start\_? (DateTime.date\_? (?=}\\ \texttt{(nextDayOfMonth (Today) 24L))))}\\ \texttt{(Event.attendees\_?}\\ \texttt{(AttendeeListHasRecipient (Execute}\\ \texttt{(refer (extensionConstraint}\\ \texttt{(RecipientWithNameLike (\textasciicircum{}(Recipient)}\\ \texttt{EmptyStructConstraint) (PersonName.apply}\\ \texttt{"David Largenstop")))))))))))}}}
\newcommand{\SmcalflowBCbrUttC}[0]{\makecell[tl]{\textit{create a meet with bob today}}}
\newcommand{\SmcalflowBCbrMeanC}[0]{\makecell[tl]{\texttt{(Yield (CreateCommitEventWrapper}\\ \texttt{(CreatePreflightEventWrapper (\&}\\ \texttt{(Event.start\_? (DateTime.date\_? (?=}\\ \texttt{(Today)))) (Event.attendees\_?}\\ \texttt{(AttendeeListHasRecipient (Execute}\\ \texttt{(refer (extensionConstraint}\\ \texttt{(RecipientWithNameLike (\textasciicircum{}(Recipient)}\\ \texttt{EmptyStructConstraint) (PersonName.apply}\\ \texttt{"bob")))))))))))}}}
\newcommand{\SmcalflowBEprUttA}[0]{\makecell[tl]{\textit{make a meeting with jeri today}}}
\newcommand{\SmcalflowBEprMeanA}[0]{\makecell[tl]{\texttt{(Yield (CreateCommitEventWrapper}\\ \texttt{(CreatePreflightEventWrapper (\&}\\ \texttt{(Event.start\_? (DateTime.date\_? (?=}\\ \texttt{(Today)))) (Event.attendees\_?}\\ \texttt{(AttendeeListHasRecipient (Execute}\\ \texttt{(refer (extensionConstraint}\\ \texttt{(RecipientWithNameLike (\textasciicircum{}(Recipient)}\\ \texttt{EmptyStructConstraint) (PersonName.apply}\\ \texttt{"jeri")))))))))))}}}
\newcommand{\SmcalflowBEprUttB}[0]{\makecell[tl]{\textit{put meeting with emlime on today}}}
\newcommand{\SmcalflowBEprMeanB}[0]{\makecell[tl]{\texttt{(Yield (CreateCommitEventWrapper}\\ \texttt{(CreatePreflightEventWrapper (\&}\\ \texttt{(Event.start\_? (DateTime.date\_? (?=}\\ \texttt{(Today)))) (Event.attendees\_?}\\ \texttt{(AttendeeListHasRecipient (Execute}\\ \texttt{(refer (extensionConstraint}\\ \texttt{(RecipientWithNameLike (\textasciicircum{}(Recipient)}\\ \texttt{EmptyStructConstraint) (PersonName.apply}\\ \texttt{"emlime")))))))))))}}}
\newcommand{\SmcalflowBEprUttC}[0]{\makecell[tl]{\textit{I want meet Dr Kennady from today}}}
\newcommand{\SmcalflowBEprMeanC}[0]{\makecell[tl]{\texttt{(Yield (CreateCommitEventWrapper}\\ \texttt{(CreatePreflightEventWrapper (\&}\\ \texttt{(Event.start\_? (DateTime.date\_? (?=}\\ \texttt{(Today)))) (Event.attendees\_?}\\ \texttt{(AttendeeListHasRecipient (Execute}\\ \texttt{(refer (extensionConstraint}\\ \texttt{(RecipientWithNameLike (\textasciicircum{}(Recipient)}\\ \texttt{EmptyStructConstraint) (PersonName.apply}\\ \texttt{"Dr Kennady")))))))))))}}}

\begin{table*}[]
\centering
\footnotesize
\scalebox{0.9}{
\begin{tabular}{cc|cl}
\multicolumn{1}{l}{}                        & \multicolumn{1}{l|}{}                                            & \multicolumn{1}{c|}{EPR}      & \multicolumn{1}{c}{CBR} \\ \hline
\multicolumn{1}{c|}{\multirow{2}{*}{\begin{tabular}[c]{@{}c@{}}Test\\ Example\end{tabular}}}  & Utterance                                                        & \multicolumn{2}{c}{\SmcalflowBGoldUtt}                             \\ \cline{2-4} 
\multicolumn{1}{c|}{}                       & \begin{tabular}[c]{@{}c@{}}Meaning\\ Representation\end{tabular} & \multicolumn{2}{c}{\SmcalflowBGoldMean}                            \\ \hline
\multicolumn{1}{c|}{\multirow{2}{*}{Top-1}} & Utterance                                                        & \multicolumn{1}{c|}{\SmcalflowBEprUttA}  & \SmcalflowBCbrUttA                 \\ \cline{2-4} 
\multicolumn{1}{c|}{}                       & \begin{tabular}[c]{@{}c@{}}Meaning\\ Representation\end{tabular} & \multicolumn{1}{c|}{\SmcalflowBEprMeanA} & \SmcalflowBCbrMeanA                \\ \hline
\multicolumn{1}{c|}{\multirow{2}{*}{Top-2}} & Utterance                                                        & \multicolumn{1}{c|}{\SmcalflowBEprUttB}  & \SmcalflowBCbrUttB                 \\ \cline{2-4} 
\multicolumn{1}{c|}{}                       & \begin{tabular}[c]{@{}c@{}}Meaning\\ Representation\end{tabular} & \multicolumn{1}{c|}{\SmcalflowBEprMeanB} & \SmcalflowBCbrMeanB                \\ \hline
\multicolumn{1}{c|}{\multirow{2}{*}{Top-3}} & \textit{Utterance}                                               & \multicolumn{1}{c|}{\SmcalflowBEprUttC}  & \SmcalflowBCbrUttC                 \\ \cline{2-4} 
\multicolumn{1}{c|}{}                       & \begin{tabular}[c]{@{}c@{}}Meaning\\ Representation\end{tabular} & \multicolumn{1}{c|}{\SmcalflowBEprMeanC} & \SmcalflowBCbrMeanC                \\ \hline
\end{tabular}
}
\caption{An example from \smcalflow{} development set where EPR is correct and CBR is incorrect along with the top-3 training examples retrieved from each retriever.}
\label{tab:smcalflowB-retrieval-examples}
\end{table*}

\newcommand{\verAutt}[0]{\makecell[tl]{\textit{which 3 seas border philippines?}}}
\newcommand{\verAmr}[0]{\makecell[tl]{\texttt{1\#) return the  philippines}\\ \texttt{2\#) return seas that border \#1}}}
\newcommand{\verButt}[0]{\makecell[tl]{\textit{what three seas surround philippines?}}}
\newcommand{\verBmr}[0]{\makecell[tl]{\texttt{1\#) return seas}\\ \texttt{2\#) return \#1 that surround the  philippines}}}
\newcommand{\verCutt}[0]{\makecell[tl]{\textit{what states does west virginia border?}}}
\newcommand{\verCmr}[0]{\makecell[tl]{\texttt{1\#) return west virginia}\\ \texttt{2\#) return border states of \#1}}}
\newcommand{\verDutt}[0]{\makecell[tl]{\textit{what states borders west virginia?}}}
\newcommand{\verDmr}[0]{\makecell[tl]{\texttt{1\#) return west virginia}\\ \texttt{2\#) return border states of \#1}}}
\newcommand{\verEutt}[0]{\makecell[tl]{\textit{which states border colorado}}}
\newcommand{\verEmr}[0]{\makecell[tl]{\texttt{1\#) return states}\\ \texttt{2\#) return \#1 that  border colorado}}}

\begin{table*}[]
\centering
\scalebox{0.9}{\footnotesize
\begin{tabular}{ll}
\toprule
Utterance & Meaning Representation \\
\midrule
 \verAutt &                \verAmr \\ \hline 
 \verButt &                \verBmr \\ \hline 
 \verCutt &                \verCmr \\ \hline 
 \verDutt &                \verDmr \\ \hline 
 \verEutt &                \verEmr \\ \bottomrule
\end{tabular}
}
\caption{Example of a cluster from the t-SNE projection of EPR on \breakdataset{}.}

\label{tab:cluster1}

\end{table*}
\newcommand{\buildAutt}[0]{\makecell[tl]{\textit{List the total scores of body builders}\\ \textit{in ascending order.}}}
\newcommand{\buildAmr}[0]{\makecell[tl]{\texttt{1\#) return body builders}\\ \texttt{2\#) return scores of  \#1}\\ \texttt{3\#) return sum of  \#2 for each  \#1}\\ \texttt{4\#) return \#3 sorted by ascending order}}}
\newcommand{\buildButt}[0]{\makecell[tl]{\textit{What are the names of body builders in}\\ \textit{descending order of total scores?}}}
\newcommand{\buildBmr}[0]{\makecell[tl]{\texttt{1\#) return body builders}\\ \texttt{2\#) return names of \#1}\\ \texttt{3\#) return scores of \#1}\\ \texttt{4\#) return sum of \#3 for each \#1}\\ \texttt{5\#) return \#2 sorted by \#4 in  descending order}}}
\newcommand{\buildCutt}[0]{\makecell[tl]{\textit{List the total points of gymnasts in}\\ \textit{descending order.}}}
\newcommand{\buildCmr}[0]{\makecell[tl]{\texttt{1\#) return gymnasts}\\ \texttt{2\#) return points of  \#1}\\ \texttt{3\#) return sum of  \#2 for each  \#1}\\ \texttt{4\#) return \#3 sorted by descending order}}}
\newcommand{\buildDutt}[0]{\makecell[tl]{\textit{What are the total points for all}\\ \textit{gymnasts, ordered by total points}\\ \textit{descending?}}}
\newcommand{\buildDmr}[0]{\makecell[tl]{\texttt{1\#) return gymnasts}\\ \texttt{2\#) return total  points  for all \#1}\\ \texttt{3\#) return \#2 ordered by total  points  descending}}}
\newcommand{\buildEutt}[0]{\makecell[tl]{\textit{List the total points of gymnasts in}\\ \textit{descending order of floor exercise}\\ \textit{points.}}}
\newcommand{\buildEmr}[0]{\makecell[tl]{\texttt{1\#) return gymnasts}\\ \texttt{2\#) return points  of  \#1}\\ \texttt{3\#) return sum of  \#2 for each  \#1}\\ \texttt{4\#) return floor exercise points  of  \#1}\\ \texttt{5\#) return \#3 sorted by \#4 in descending order}}}

\begin{table*}[]
\centering
\scalebox{0.9}{\footnotesize
\begin{tabular}{ll}
\toprule
 Utterance & Meaning Representation \\
\midrule
\buildAutt &              \buildAmr \\ \hline 
\buildButt &              \buildBmr \\ \hline 
\buildCutt &              \buildCmr \\ \hline 
\buildDutt &              \buildDmr \\ \hline 
\buildEutt &              \buildEmr \\ \bottomrule
\end{tabular}
}
\caption{Example of a cluster from the t-SNE projection of EPR on \breakdataset{}.}

\label{tab:cluster2}

\end{table*}
\newcommand{\adsaAutt}[0]{\makecell[tl]{\textit{Show the locations that have both}\\ \textit{performances with more than 2000}\\ \textit{attendees and performances with less}\\ \textit{than 1000 attendees.}}}
\newcommand{\adsaAmr}[0]{\makecell[tl]{\texttt{1\#) return performances}\\ \texttt{2\#) return attendees of \#1}\\ \texttt{3\#) return the  number of  \#2 for each \#1}\\ \texttt{4\#) return \#1 where \#3 is  more than  2000}\\ \texttt{5\#) return \#1 where \#3 is  less than  1000}\\ \texttt{6\#) return the  locations of \#4}\\ \texttt{7\#) return the  locations of \#5}\\ \texttt{8\#) return the  locations in both \#6 and \#7}}}
\newcommand{\adsaButt}[0]{\makecell[tl]{\textit{Show the theme for exhibitions with both}\\ \textit{records of an attendance below 100 and}\\ \textit{above 500.}}}
\newcommand{\adsaBmr}[0]{\makecell[tl]{\texttt{1\#) return exhibitions}\\ \texttt{2\#) return attendances of \#1}\\ \texttt{3\#) return number of  \#2 for each \#1}\\ \texttt{4\#) return \#1 where \#3 is  below 100}\\ \texttt{5\#) return \#1 where \#3 is  above 500}\\ \texttt{6\#) return \#1 of both \#4 and \#5}\\ \texttt{7\#) return themes for \#6}}}
\newcommand{\adsaCutt}[0]{\makecell[tl]{\textit{Which themes have had corresponding}\\ \textit{exhibitions that have had attendance}\\ \textit{both below 100 and above 500?}}}
\newcommand{\adsaCmr}[0]{\makecell[tl]{\texttt{1\#) return themes}\\ \texttt{2\#) return exhibitions with \#1}\\ \texttt{3\#) return attendances of  \#2}\\ \texttt{4\#) return \#1 where \#3 is  lower than 100}\\ \texttt{5\#) return \#1 where \#3 is  higher than 500}\\ \texttt{6\#) return \#1 of  both \#4 and  \#5}}}
\newcommand{\adsaDutt}[0]{\makecell[tl]{\textit{Show the publishers that have}\\ \textit{publications with price higher than}\\ \textit{10000000 and publications with price}\\ \textit{lower than 5000000.}}}
\newcommand{\adsaDmr}[0]{\makecell[tl]{\texttt{1\#) return publishers}\\ \texttt{2\#) return publications of \#1}\\ \texttt{3\#) return prices of \#2}\\ \texttt{4\#) return \#1 where \#3 is  higher than 10000000}\\ \texttt{5\#) return \#1 where \#3 is  lower than 5000000}\\ \texttt{6\#) return \#1 of both \#4 and \#5}}}
\newcommand{\adsaEutt}[0]{\makecell[tl]{\textit{Show the famous titles of the artists}\\ \textit{with both volumes that lasted more than}\\ \textit{2 weeks on top and volumes that lasted}\\ \textit{less than 2 weeks on top.}}}
\newcommand{\adsaEmr}[0]{\makecell[tl]{\texttt{1\#) return artists}\\ \texttt{2\#) return volumes  of \#1}\\ \texttt{3\#) return weeks  on top that  \#2 lasted}\\ \texttt{4\#) return number of  \#3 for each \#2}\\ \texttt{5\#) return \#1 where \#4 is  more than  2}\\ \texttt{6\#) return \#1 where \#4 is  less than  2}\\ \texttt{7\#) return \#1 in both \#5 and \#6}\\ \texttt{8\#) return famous titles of \#7}}}

\begin{table*}[]
\centering
\scalebox{0.9}{\footnotesize
\begin{tabular}{ll}
\toprule
Utterance & Meaning Representation \\
\midrule
\adsaAutt &               \adsaAmr \\ \hline 
\adsaButt &               \adsaBmr \\ \hline 
\adsaCutt &               \adsaCmr \\ \hline 
\adsaDutt &               \adsaDmr \\ \hline 
\adsaEutt &               \adsaEmr \\ \bottomrule
\end{tabular}
}
\caption{Example of a cluster from the t-SNE projection of EPR on \breakdataset{}.}

\label{tab:cluster3}

\end{table*}
\newcommand{\metalAutt}[0]{\makecell[tl]{\textit{What is the metal thing next to the}\\ \textit{small cylinder?}}}
\newcommand{\metalAmr}[0]{\makecell[tl]{\texttt{1\#) return the  small cylinder}\\ \texttt{2\#) return things}\\ \texttt{3\#) return \#2 that are  metal}\\ \texttt{4\#) return \#3 that are  next to  \#1}}}
\newcommand{\metalButt}[0]{\makecell[tl]{\textit{What is the purple thing next to the}\\ \textit{brown thing?}}}
\newcommand{\metalBmr}[0]{\makecell[tl]{\texttt{1\#) return the  brown thing}\\ \texttt{2\#) return things}\\ \texttt{3\#) return \#2 that  are purple}\\ \texttt{4\#) return \#3 that  are next to  \#1}}}
\newcommand{\metalCutt}[0]{\makecell[tl]{\textit{What is the gray thing next to the}\\ \textit{block?}}}
\newcommand{\metalCmr}[0]{\makecell[tl]{\texttt{1\#) return gray thing}\\ \texttt{2\#) return the  block}\\ \texttt{3\#) return \#1 next to \#2}}}
\newcommand{\metalDutt}[0]{\makecell[tl]{\textit{What is the shiny thing next to the}\\ \textit{cylinder?}}}
\newcommand{\metalDmr}[0]{\makecell[tl]{\texttt{1\#) return shiny thing}\\ \texttt{2\#) return cylinder}\\ \texttt{3\#) return \#1 next to \#2}}}
\newcommand{\metalEutt}[0]{\makecell[tl]{\textit{What is the thing in front of the red}\\ \textit{square?}}}
\newcommand{\metalEmr}[0]{\makecell[tl]{\texttt{1\#) return things}\\ \texttt{2\#) return squares}\\ \texttt{3\#) return \#2 that is red}\\ \texttt{4\#) return \#1 that is in  front of \#3}}}

\begin{table*}[]
\centering
\scalebox{0.9}{\footnotesize
\begin{tabular}{ll}
\toprule
 Utterance & Meaning Representation \\
\midrule
\metalAutt &              \metalAmr \\ \hline 
\metalButt &              \metalBmr \\ \hline 
\metalCutt &              \metalCmr \\ \hline 
\metalDutt &              \metalDmr \\ \hline 
\metalEutt &              \metalEmr \\ \bottomrule
\end{tabular}

}
\caption{Example of a cluster from the t-SNE projection of EPR on \breakdataset{}.}

\label{tab:cluster4}

\end{table*}
\newcommand{\purpleAutt}[0]{\makecell[tl]{\textit{Is the purple thing behind the big red}\\ \textit{thing?}}}
\newcommand{\purpleAmr}[0]{\makecell[tl]{\texttt{1\#) return purple thing}\\ \texttt{2\#) return big red thing}\\ \texttt{3\#) return Is \#1 behind \#2}}}
\newcommand{\purpleButt}[0]{\makecell[tl]{\textit{is the purple sphere in front of the}\\ \textit{blue cube?}}}
\newcommand{\purpleBmr}[0]{\makecell[tl]{\texttt{1\#) return the  purple sphere}\\ \texttt{2\#) return the  blue cube}\\ \texttt{3\#) return if  \#1 is  in front of \#2}}}
\newcommand{\purpleCutt}[0]{\makecell[tl]{\textit{is the gray sphere behind the green}\\ \textit{cylinder?}}}
\newcommand{\purpleCmr}[0]{\makecell[tl]{\texttt{1\#) return the  green cylinder}\\ \texttt{2\#) return the  gray sphere}\\ \texttt{3\#) return if  \#2 is  behind \#1}}}
\newcommand{\purpleDutt}[0]{\makecell[tl]{\textit{is the red cube in front of the yellow}\\ \textit{ball?}}}
\newcommand{\purpleDmr}[0]{\makecell[tl]{\texttt{1\#) return the  red cube}\\ \texttt{2\#) return the  yellow ball}\\ \texttt{3\#) return if \#1 is in  front of \#2}}}
\newcommand{\purpleEutt}[0]{\makecell[tl]{\textit{Is the blue ball in front of the silver}\\ \textit{cube?}}}
\newcommand{\purpleEmr}[0]{\makecell[tl]{\texttt{1\#) return blue ball}\\ \texttt{2\#) return silver cube}\\ \texttt{3\#) return is  \#1 in front of \#2}}}

\begin{table*}[]
\centering
\scalebox{0.9}{\footnotesize
\begin{tabular}{ll}
\toprule
  Utterance & Meaning Representation \\
\midrule
\purpleAutt &             \purpleAmr \\ \hline 
\purpleButt &             \purpleBmr \\ \hline 
\purpleCutt &             \purpleCmr \\ \hline 
\purpleDutt &             \purpleDmr \\ \hline 
\purpleEutt &             \purpleEmr \\ \bottomrule
\end{tabular}

}
\caption{Example of a cluster from the t-SNE projection of EPR on \breakdataset{}.}

\label{tab:cluster5}

\end{table*}

\end{document}